\documentclass{article}

\usepackage{microtype}
\usepackage{graphicx}
\usepackage{subfigure}
\usepackage{booktabs} 

\usepackage{hyperref}

\usepackage{csquotes}
\usepackage{multirow, makecell}
\usepackage{float}

\usepackage{algorithm}
\usepackage{algpseudocode}  

\renewcommand{\algorithmicrequire}{\textbf{Input:}}  
\renewcommand{\algorithmicensure}{\textbf{Output:}} 

\usepackage[accepted]{icml2024}

\usepackage{amsmath}
\usepackage{amssymb}
\usepackage{mathtools}
\usepackage{amsthm}

\usepackage[capitalize,noabbrev]{cleveref}

\theoremstyle{plain}
\newtheorem{theorem}{Theorem}[section]
\newtheorem{proposition}[theorem]{Proposition}

\theoremstyle{definition}

\theoremstyle{remark}

\usepackage[textsize=tiny]{todonotes}

\allowdisplaybreaks[4]
\icmltitlerunning{Keypoint-based Progressive Chain-of-Thought Distillation for LLMs}

\begin{document}

\twocolumn[
\icmltitle{Keypoint-based Progressive Chain-of-Thought Distillation for LLMs}




\begin{icmlauthorlist}
\icmlauthor{Kaituo Feng}{bit}
\icmlauthor{Changsheng Li}{bit}
\icmlauthor{Xiaolu Zhang}{ant}
\icmlauthor{Jun Zhou}{ant}
\icmlauthor{Ye Yuan}{bit}
\icmlauthor{Guoren Wang}{bit,hebei}
\end{icmlauthorlist}

\icmlaffiliation{bit}{Beijing Institute of Technology}
\icmlaffiliation{ant}{Ant Group}
\icmlaffiliation{hebei}{Hebei Province Key Laboratory of Big Data Science and Intelligent Technology}

\icmlcorrespondingauthor{Changsheng Li}{lcs@bit.edu.cn}

\icmlkeywords{Machine Learning, ICML}

\vskip 0.3in
]



\printAffiliationsAndNotice{}  

\begin{abstract}
 Chain-of-thought distillation  is a powerful technique for transferring reasoning abilities from large language models (LLMs) to smaller student models. 
Previous methods typically require the student to mimic the step-by-step rationale produced by LLMs, often facing the following challenges:
(i) Tokens within a rationale vary in significance, and treating them equally may fail to accurately mimic keypoint tokens, leading to reasoning errors.
(ii) They usually distill knowledge by consistently predicting all the steps in a rationale, which  falls short in distinguishing the learning order of step generation.
This diverges from the human cognitive progression of starting with easy tasks and advancing to harder ones, resulting in sub-optimal outcomes.
To this end, we propose a unified framework, called KPOD, to address these issues. 
Specifically, we propose a token weighting module utilizing mask learning to encourage accurate mimicry of keypoint tokens by the student during distillation.
Besides, we develop an in-rationale progressive distillation strategy, starting with training the student  to generate the final reasoning steps and gradually extending to cover the entire rationale.
To accomplish this, a weighted token generation loss is proposed to assess step reasoning difficulty, and a value function is devised to schedule the progressive distillation  by considering both step difficulty and question diversity.
Extensive experiments on four reasoning benchmarks illustrate our KPOD  outperforms previous methods by a large margin.
\end{abstract}

\section{Introduction}

Large language models (LLMs) have demonstrated remarkable reasoning capabilities via chain-of-thought (CoT) prompting (e.g., \enquote{Let's think step-by-step}), which prompts LLMs to generate a step-by-step rationale to help reasoning \cite{kojima2022large, wei2022chain}. 
However, such abilities usually emerge in extremely large models, especially those with over 100 billion parameters
\cite{fu2023specializing,hoffmann2022training}
, such as 175B GPT-3 \cite{brown2020language} and 540B PaLM \cite{chowdhery2023palm}. The substantial amount of parameters unavoidably leads to high inference costs and makes it challenging to deploy LLMs in environments with limited computational resources \cite{hsieh2023distilling}. To tackle with this, a recent surge of works, known as CoT distillation, has arisen as a promising avenue to distill reasoning capabilities of LLMs to smaller student models \cite{lietal2023symbolic,wangetal2023scott,fu2023specializing}. The core idea of  these methods is to require the student model to mimic  the step-by-step rationale generated by LLMs in response to a question.

However, current CoT distillation methods often encounter the following two issues:
First, in a rationale, each token carries different levels of importance in the reasoning process. 
Certain keypoint tokens play a pivotal role in reasoning, while other tokens are of less importance or even irrelevant to the reasoning process.
For instance, consider a step in a rationale: \enquote{\textit{Next, we just need to simply add up the calories from the lettuce and cucumber: 30 + 80 = 110}}. Here, terms like \enquote{\textit{just}}, \enquote{\textit{simply}}  are reasoning-irrelevant, whereas the calculation \enquote{\textit{30 + 80 = 110}} stands out as the keypoint for reasoning. 
The reasoning-irrelevant tokens can be replaced without negative effects, but even a slight deviation from the keypoint token could result in errors in reasoning.
Therefore, it's crucial for the student model to focus on the precise mimicry of these keypoint tokens. Nevertheless, previous CoT distillation methods usually treat all tokens equally during distillation \cite{lietal2023symbolic,wangetal2023scott}.

The second issue stems from the fact that previous approaches usually demand the student model to consistently learn all the steps in a rationale throughout the distillation process, without distinguishing the learning order of step generation.
This distillation strategy diverges from the human cognitive pattern that progresses from easier tasks to more challenging ones.
This deviation might lead to sub-optimal outcomes.
In the process of human or biological agent learning, ability acquisition doesn't simply stem from random tasks \cite{molina1998modulation}.
Instead,  there is an organized progression from easy tasks to hard tasks for them to acquire capabilities, especially for complex skills such as reasoning \cite{peterson2004day,krueger2009flexible,benoit2013young}. In the field of machine learning, this ordered learning paradigm is regarded as curriculum learning \cite{bengio2009curriculum}. Inspired by this, we intend to develop a progressive CoT distillation strategy to facilitate the student model acquire reasoning ability from easy to hard. However, directly applying previous curriculum learning strategies to CoT distillation could be inferior because of the following two reasons: (i) 
They overlook the step-by-step reasoning nature where each reasoning step within a rationale may possess varying reasoning difficulty, resulting in sub-optimal difficulty assessment.
(ii) As aforementioned,  a step in the rationale might contain many tokens that are not crucial to the reasoning process. When assessing  the difficulty of step generation, it may be dominated by these inessential tokens, thereby inaccurately reflecting the challenge of obtaining the expected outcome for a reasoning step.

In this paper, we propose Keypoint-based Progressive CoT Distillation for LLMs dubbed KPOD, with the goal of addressing the above two issues in a unified framework. First, we propose a rationale token weighting module to determine the token significance for distillation. It learns to generate masks for inessential tokens to the reasoning process via two distinctive loss functions: An answer prediction loss is introduced to encourage the module to utilize the question with the masked rationale to derive the answer, while a mask ratio loss is designed to maximize the ratio of masked tokens in the rationale. By doing so, the obtained probability of not masking a token can serve as an indicator of its significance weight.
 Second,  we develop an in-rationale progressive distillation strategy that orders the learning sequence from easy reasoning to hard reasoning within the rationale of a question. 
This strategy begins by training the student model to generate the last few reasoning steps of the rationale, given the question with preceding steps of this rationale as input. Subsequently, it progressively extends to generate the entire rationale using only the question as input. 
To precisely assess each step's reasoning difficulty, we propose a token generation loss based on the derived token significance,  aiming to eliminate the negative effects of reasoning-irrelevant tokens.
Finally, we design a value function to dynamically determine the number of steps  taken as input at each stage, thereby automatically adjusting their learning difficulty. 
Meanwhile, we leverage the value function to select diverse questions, so as to prevent over-fitting \cite{{jiang2014self,liang2021token}}.



Our contributions can be summarized as: 
1) We propose a general and principled framework for CoT distillation, which simultaneously considers token significance and reasoning difficulty within a rationale during distillation. 
2) We design a rationale token weighting module through mask learning to determine the token significance for reasoning. This allows the student  to concentrate more on keypoint tokens.
3) We devise an in-rationale progressive CoT distillation strategy to schedule the  learning order of reasoning steps within a rationale.
This enables the student to progressively acquire reasoning abilities in an easy-to-hard manner.
4) Extensive experiments on four reasoning benchmarks validate the effectiveness of our KPOD, showcasing significant performance improvements compared to baselines.

\section{Related Works}

\textbf{Chain-of-Thought Reasoning.}
The concept of employing step-by-step language rationales to aid in solving reasoning problems can be traced back to pioneering works \cite{ling2017program}. Inspired by this, chain-of-thought prompting \cite{wei2022chain} has been proposed to enable LLMs to generate intermediate
reasoning steps that contribute to the final answer via few-shot CoT demonstrations. This prompting approach has illustrated remarkable performance gain for LLMs in reasoning related tasks \cite{zhang2022automatic,wang2023cue}. In addition, researchers find that LLMs can also obtain impressive reasoning performance by zero-shot CoT \cite{kojima2022large} without task-related demonstrations. This is achieved by only using a single sentence \enquote{Let’s think step by step} for prompting. Recently, a number of CoT prompting methods have demonstrated effectiveness in enhancing the reasoning performance of LLMs \cite{diao2023active,yang2023large}, such as SC-CoT \cite{wang2022self}, Auto-CoT \cite{zhang2022automatic}, Multimodal-CoT \cite{zhang2023multimodal}, etc. 
However, the emergence of CoT reasoning capabilities in LLMs typically requires models with more than 100 billion parameters \cite{wei2022chain,fu2023specializing}, making it resource-consuming for deployment.

\textbf{CoT Distillation.}
Knowledge distillation has been widely studied for model compression across various fields \cite{magisteretal2023teaching, feng2024road}.
Recently, CoT Distillation has emerged as a promising avenue to transfer the step-by-step reasoning capabilities of LLMs to smaller student models \cite{hsieh2023distilling,hoetal2023large}. The key idea of CoT distillation is to make the student model mimic the step-by-step rationale generated by LLMs in response to a question. In this context, the rationale can be interpreted as the LLMs' explanation of how to derive the final answer of a question, akin to the soft label used in conventional knowledge distillation \cite{hinton2015distilling,feng2022freekd}. The representative works of CoT distillation include: SCoTD \cite{lietal2023symbolic} introduces a symbolic CoT distillation method that enables smaller models to self-rationalize for reasoning via learning rationales from LLMs. 
Specialized KD \cite{fu2023specializing} is proposed to train a small language model specialized for reasoning in four distinct in-context scenarios.
MCC-KD \cite{chenetal2023mcc} adopts diverse rationales for distillation and attempts to ensure their consistency.
SCOTT \cite{wangetal2023scott} designs a faithful CoT distillation strategy to 
make the student reason faithfully via counterfactual training. However, these methods fail to consider the reasonable learning order of the reasoning steps within a rationale, leading to sub-optimal performance.

\textbf{Curriculum Learning.} Early researches in cognitive science emphasize the significance of the easy-to-hard learning pattern to acquire knowledge \cite{elman1993learning}. Inspired by this, the pioneer work \cite{bengio2009curriculum} introduces the concept of curriculum learning (CL) to the machine learning field by gradually including samples from easy to hard for training. In recent years, a variety of CL methods have been proposed to enhance the model performance \cite{kong2021adaptive,wang2021curgraph}. 
For instance, Adaptive CL \cite{kong2021adaptive} proposes to utilize the loss of the model to dynamically adjust the difficulty score of each sample.
SPL \cite{wanetal2020self} introduces the curriculum learning to the neural machine translation domain via introducing the token-level and sentence-level confidence score. ICL \cite{jiaetal2023sample} devises a curriculum learning method that organizes the curriculum within the token sequence of a sample for natural language generation tasks.
However, as aforementioned,  applying these CL methods directly to CoT distillation could yield inferior performance.


\section{Proposed Method}

\begin{figure*}
  \centering
  \includegraphics[width=1.0\linewidth]{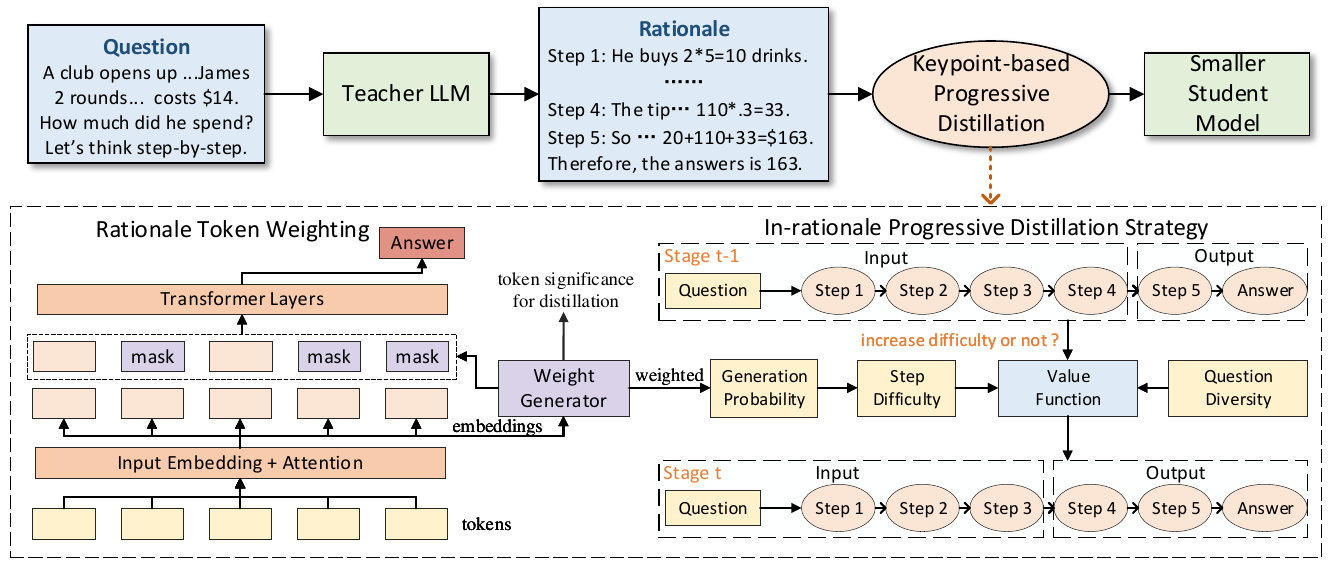}
    \vspace{-0.2in}
  \caption{
    An illustration of our KPOD framework. KPOD first determines the keypoint tokens for distillation through designing a rationale token weighting module based on mask learning. Then, an in-rationale progressive distillation strategy is devised to organize the learning order within rationale, so as to enable the student to acquire the reasoning capabilities in an easy-to-hard manner.
  }
  \label{main}
  \vspace{-0.15in}
\end{figure*}

\subsection{Preliminaries and Problem Setting}

The goal of CoT distillation is to transfer the reasoning capability of large language models (LLMs) to smaller student models via distilling the rationales produced by LLMs. 
We denote the dataset as $\mathcal{D} = \{(x^{(i)}, y^{(i)})\}$, where $x^{(i)}$ is the $i$-th reasoning question and $y^{(i)}$ is the corresponding answer. Following previous CoT distillation works \cite{hoetal2023large,chenetal2023mcc} , we adopt zero-shot CoT \cite{kojima2022large} to prompt the teacher LLMs to generate step-by-step rationale $r^{(i)}$ for each question $x^{(i)}$. 
The reasoning template takes the following format: \enquote{\textit{Q: \textless{}$x^{(i)}$\textgreater{} A: \textless{}p\textgreater{} \textless{}$r^{(i)}$\textgreater{} Therefore, the answer is \textless{}$y^{(i)}$\textgreater{}}}, 
where \textless{}$p$\textgreater{} is the zero-shot CoT prompt such as \enquote{Let’s think step by step}.
Then, the student is trained to generate the concatenated sequence of rationale tokens $r^{(i)}$ and answer tokens $y^{(i)}$,
given the question $x^{(i)}$ as input. 
The standard negative log-likelihood loss for training the student model can be formulated as:
\begin{align} \nonumber
\label{gen}
\setlength{\abovedisplayskip}{0.01in}
\setlength{\belowdisplayskip}{0.01in}
    \mathcal{L}\! &=-\!\sum_i\sum_j \log\! P(r_j^{(i)} | r_{<j}^{(i)}, x^{(i)};\theta^s) \\
    &\ \ \ \ -\sum_i\sum_j \log\! P(y_j^{(i)} | y_{<j}^{(i)},r^{(i)},x^{(i)};\theta^s),
\end{align}
where $r_j^{(i)}$ and $y_j^{(i)}$ represent the $j$-th token in the rationale sequence $r^{(i)}$ and the answer sequence $y^{(i)}$, respectively. 
$\theta^s$ denotes the parameters of the student model. The first term of Eq.(\ref{gen}) enables the student to mimic the rationale produced by LLMs, while the second term  aims to train the student to output the final answer based on the rationale.
By minimizing this loss, the student model can learn to generate the step-by-step rationale for  deriving the final answer.

\subsection{Framework Overview}
As aforementioned, there are two key issues for CoT distillation methods: (i) 
Equally treating each token for distillation may  make the student fail to mimic keypoint tokens accurately, leading to reasoning errors. 
(ii) Distilling the steps within a rationale without explicitly considering the learning order of step generation might lead to sub-optimal outcomes.
To tackle these two issues, we propose a new CoT distillation framework KPOD, as illustrated in Figure \ref{main}. Our framework mainly consists of two components: a rationale token weighting component based on mask learning is proposed to determine the token significance  for distillation. This encourages the student to faithfully replicate the crucial keypoint tokens; A progressive distillation component within the rationale is designed to establish a structured learning order for the reasoning steps. This guides the student model to progressively develop its reasoning abilities from simpler to more complex tasks, aligning with the proficiency of teacher LLMs. 
It's worth noting that the obtained token significance weight fulfills two distinct functions in our framework: firstly, it encourages precise mimicry of keypoint tokens during distillation, and secondly, it mitigates the negative effects of inessential tokens when assessing step difficulty.
Next, we will primarily delve into the detailed introduction of the two components in our framework.

\subsection{Rationale Token Weighting}


In this section, we introduce our rationale token weighting module, which determines the significance of each token via learning to mask reasoning-irrelevant token.

\textbf{Weight Generation.} 
First, we intend to generate distinct significance weights for different tokens by leveraging their embeddings. This facilitates the estimation of their importance according to their characteristics. 
To achieve this, we feed the rationale tokens into a pre-trained input embedding layer, followed by a self-attention layer to encode in-context information. This process is formulated as:
\begin{equation}
\setlength{\abovedisplayskip}{0.05in}
\setlength{\belowdisplayskip}{0.05in}
    e^{(i)} = \mathrm{Att}(\mathrm{Emb}(r^{(i)})),
\end{equation}
where $\mathrm{Emb}$ and $\mathrm{Att}$ denote the input embedding layer and the self-attention layer, respectively. $e^{(i)}$ is the embedding matrix containing the embeddings for each token in $r^{(i)}$. Subsequently, the embedding $e_j^{(i)}$ of each token is fed into a weight generator, producing the significance weight as:
\begin{equation}
\setlength{\abovedisplayskip}{0.05in}
\setlength{\belowdisplayskip}{0.05in}
    w_j^{(i)} = \sigma(f_w (e_j^{(i)})),
\end{equation}
where $w_j^{(i)}$ is the probability of the $j$-th token not being masked, serving as an indicator of its significance level. $f_w$ is the weight generator and $\sigma$ is the sigmoid activation function \cite{narayan1997generalized}. In this paper, we employ a simple two-layer MLP as the weight generator.

\textbf{Reasoning-irrelevant Mask Learning.} 
To optimize the weight generator, we formulate two loss functions: an answer prediction loss that encourages the module to utilize the question with the masked rationale for answer derivation, and a mask ratio loss aiming to maximize the ratio of masked tokens in the rationale.
This allows the weight generator to generate low values of $w_j^{(i)}$ for tokens irrelevant to reasoning and high values for keypoint tokens.
Next, we will introduce these two losses in detail.

Firstly, considering that sampling the discrete mask policy $m_j^{(i)}\in \{0,1\}$ from the distribution of $w_j^{(i)}$ is non-differentiable, we adopt the Gumbel-Softmax sampling \cite{jang2016categorical} to avoid this issue:
\begin{equation}
\setlength{\abovedisplayskip}{0.1in}
\setlength{\belowdisplayskip}{0.1in}
    m_j^{(i)} = \mathrm{GumbelSoftmax}(w_j^{(i)}), 
\end{equation}
where $\mathrm{GumbelSoftmax}$ represents the Gumbel-Softmax sampling \cite{jang2016categorical}. $m_j^{(i)}=0$ denotes that the $j$-th token is masked, while $m_j^{(i)}=1$ denotes that the $j$-th token is not masked. By applying the mask $m_j^{(i)}$ to each token $r^{(i)}_j$ in rationale $r^{(i)}$, we can obtain the masked rationale, denoted as ${r[m]}^{(i)}$.

Then, we input ${r[m]}^{(i)}$ into the transformer layers, with the goal of  obtaining the correct answer by using the masked rationale. Here we initialize the transformer layers using the pre-trained FlanT5-Large \cite{chung2022scaling}. Considering that certain steps of the rationale sometimes contain the reasoning results of previous steps, shortcuts may be taken for the answer prediction via neglecting previous steps. To eliminate this phenomenon, we expect the transformer to predict the answer based on the question with any prefix of the masked rationale.
The answer prediction loss $\mathcal{L}_p$ for question $x^{(i)}$ can be written as:
\begin{equation}
\setlength{\abovedisplayskip}{0.05in}
\setlength{\belowdisplayskip}{0.05in}
    \mathcal{L}_p = -\sum_k \sum_j \log P(y_j^{(i)} | y_{<j}^{(i)}, {r[m]}^{(i)}_{<k}, x^{(i)};\theta^w),
\end{equation}
where $y^{(i)}$ is the answer and $x^{(i)}$ is the question. $\theta^w$ represents the parameters of this rationale token weighting module. ${r[m]}^{(i)}_{<k}$ represents the preceding $k$ tokens of the masked rationale ${r[m]}^{(i)}$. 
By optimizing $\mathcal{L}_p$, the transformer can be used to predict the answer by taking as input the question and the prefix of the masked rationale.
Meanwhile, the weight generator is encouraged to generate large weights for the keypoint tokens, preventing them from being masked to facilitate the answer prediction.

Moreover, to eliminate redundant tokens for reasoning, a mask ratio loss $\mathcal{L}_m$ is presented as:
\begin{equation}
\setlength{\abovedisplayskip}{0.05in}
\setlength{\belowdisplayskip}{0.05in}
    \mathcal{L}_m = \sum_j m_j^{(i)}.
\end{equation}
By optimizing $\mathcal{L}_m$, we enable the weight generator to identify the insignificant tokens in the reasoning process and generate lower weights for them.

Finally, the overall loss function for training this module can be expressed as:
\begin{equation}
\label{weighting}
\setlength{\abovedisplayskip}{0.05in}
\setlength{\belowdisplayskip}{0.05in}
    \mathcal{L}_k = \mathcal{L}_p  + \alpha \mathcal{L}_m,
\end{equation}
where $\alpha$ is a balancing hyper-parameter. By optimizing $\mathcal{L}_k$, we can achieve the goal of determining the significance weight $w_j^{(i)}$ for each token within a rationale. 


\subsection{In-rationale Progressive Distillation}

In this section, we elaborate our proposed in-rationale progressive distillation strategy, which schedules the learning order within a rationale.

\textbf{Step Difficulty Assessment.} Firstly, we assess the difficulty of  each reasoning step in the rationale, so as to facilitate the learning order scheduling. In this work, we utilize the symbol \enquote{.} to separate steps in a rationale. As mentioned above, there could exist many reasoning-irrelevant tokens,
and it is crucial to ensure that the difficulty evaluation is not influenced by them.
Therefore, we propose a weighted token generation loss to calculate the difficulty value $d_k^{(i)}$ of the $k$-th reasoning step  in the rationale $r^{(i)}$ as:
\begin{equation}
\label{diff}
\setlength{\abovedisplayskip}{0.05in}
\setlength{\belowdisplayskip}{0.05in}
    d_k^{(i)} = - \sum_{j=p_k}^{q_k} \hat{w}_j^{(i)} \log P(r_j^{(i)} | r_{<j}^{(i)}, x^{(i)}; \theta^s),
\end{equation}
where $p_k$ and $q_k$ denote the start position and end position of the $k$-th step in the rationale, respectively. Here we directly use the pre-trained student model $\theta^s$ (e.g., LLaMA-7B \cite{touvron2023llama}) before distillation to evaluate the generation probability $P(r_j^{(i)} | r_{<j}^{(i)}, x^{(i)};\theta^s)$. $\hat{w}_j^{(i)} = \mathrm{softmax}(w_j^{(i)})$ represents the significance weight normalized by softmax \cite{bridle1989training} within the token weights in the $k$-th step. In this way, the obtained step difficulty can be more concentrated on the difficulty of generating keypoint tokens, providing a more faithful reflection of the difficulty in deriving the correct outcome of each reasoning step.

\textbf{Progressive Distillation.} Based on the step difficulty scores, we devise an in-rationale progressive distillation strategy to guide the student model learning each rationale in an easy-to-hard fashion. This strategy initiates with training the student model to generate the final few reasoning steps of the rationale using previous steps combined with the question as input, and progressively expands to produce the complete rationales.
Supposed that we schedule the student model to output the last $n_i-c_i(t)$ steps of the $i$-th rationale at stage $t$, The difficulty $h_i(S(t))$ of generating these steps can be formulated as:
\begin{equation}
\setlength{\abovedisplayskip}{0.05in}
\setlength{\belowdisplayskip}{0.05in}
    h_i (S(t)) = \sum_{j=c_i(t)+1}^{n_i} d_j^{(i)}, 
\end{equation}
where $n_i$ is the total number of steps in the $i$-th rationale $r^{(i)}$ and $c_i(t)$ is the scheduled number of input steps of $r^{(i)}$ at stage $t$. $d_j^{(i)}$ is the difficulty of the $j$-th step in $r^{(i)}$. $S(t)$ is used to decide the value of $c_i(t)$ at stage $t$, which will be introduced later. In this paper, we treat each training epoch as a stage.

To facilitate selecting diverse questions to increase difficulty at each stage, we configure an overall learning difficulty $D(t)$ for stage $t$ rather than a hard threshold for each question. This means that the difficulty sum of all questions should not exceed $D(t)$ at stage $t$. We set the growth rate of $D(t)$ to be $\frac{dD(t)}{dt} = ut^p$, where $p>0$ and $u>0$ are the parameters to control the growth rate. By integrating the growth rate with respect to $t$, we can derive $D(t)$ as:
\begin{equation}
\setlength{\abovedisplayskip}{0.05in}
\setlength{\belowdisplayskip}{0.05in}
    D(t) = \frac{ut^{p+1}}{p+1} + C_0,
\end{equation}
where $C_0$ represents the initial overall learning difficulty at stage $0$. By letting $D(t)$ achieve the maximum difficulty $B$ of the dataset at stage $T$: $D(T)=B=\sum_i \sum_{j=1}^{n_i} d_j^{(i)}$, we can derive $u = \frac{(B-C_0)(p+1)}{T^{p+1}}$, where $p$ and $C_0$ are the pre-defined hyper-parameters.

When entering stage $t$ from stage $t-1$, it's required to select a set of  questions to increase difficulty. We achieve this by reducing a number of input steps $\Delta_s$ for the selected questions as:
\begin{equation}
\label{ci}
\setlength{\abovedisplayskip}{0.05in}
\setlength{\belowdisplayskip}{0.05in}
    c_i (t) = c_i(t-1) - q_i(t) \cdot \Delta_s,\ \ s.t. \Delta H(S(t)) \leq \Delta D(t),  
\end{equation}
where  $c_i(t)$ is the scheduled number of input steps of the $i$-th question at stage $t$. 
Let $S(t)$ denote the selected question set for increasing difficulty at stage $t$. Then, $q_i(t) \in \{0,1\}$ represents whether $i$ belongs to $S(t)$.
If $i \in S(t)$, then $q_i(t)=1$; otherwise, $q_i(t)=0$.
$\Delta_s$ is the pre-defined number for reducing input steps. $\Delta H(S(t)) = \sum_i h_i(S(t)) -\sum_i h_i(S(t-1))$ is the sum of the increased difficulty and $\Delta D(t) = D(t) - \sum_i h_i(S(t-1)) $ is the ceiling magnitude for the increased difficulty.  

Then, in order to determine whether a question should increase difficulty, we design a value function $F$. The goal of this value function is two-fold: One is to align the increased difficulty as closely as possible with the defined magnitude, and the other is to ensure a diverse set of questions for escalating difficulty to prevent overfitting \cite{jiang2014self}. The value function $F$ is designed as:
\begin{equation}
\setlength{\abovedisplayskip}{0.05in}
\setlength{\belowdisplayskip}{0.05in}
    F(S(t)) = -(\Delta D(t)\! -\! \Delta H(S(t))) \!+\! \beta\! \sum_{k=1}^K\! \sqrt{|C_k \!\cap\! S(t)|},
\end{equation}
where $\beta$ is a trade-off hyper-parameter. The first term measures the closeness of $\Delta H(S(t))$ to $\Delta D(t)$ and the second term measures the diversity of selected question set based on clustering. Specifically, $C_k$ is the question set of the $k$-th cluster and $K$ is the number of clusters. In this paper, we conduct K-means clustering \cite{bradley2000constrained} to cluster the question based on its embedding, which is calculated by the average of the GloVe \cite{pennington2014glove} word embedding. $S(t)$ is the selected question set. By using the square root operation, our aim is to promote a balanced distribution of questions within each cluster in the selected question set. This approach ensures that the diversity of the chosen question set is maintained.

The optimization of $F(S(t))$ can be formulated as:
\begin{equation}
\setlength{\abovedisplayskip}{0.05in}
\setlength{\belowdisplayskip}{0.05in}
\label{max}
    \max_{S(t)} F(S(t)),\ \  s.t. \Delta H(S(t)) \leq \Delta D(t).
\end{equation}
By maximizing $F(S(t))$, we can achieve the goal of selecting diverse questions to increase difficulty with close proximity to $\Delta D(t)$. However, this is a combination optimization problem subject to the knapsack constraint,  and solving it is known to be NP-hard. 
Fortunately, we can prove that $F(S(t))$ satisfies the condition of monotone and submodular. Therefore, it can be approximately solved by a submodular maximization algorithm FTGP \cite{li2022submodular} in linear time with an approximation ratio guarantee, as formulated in Proposition \ref{prop}. The proof of Proposition \ref{prop} can be found in Appendix \ref{proof}. 

\begin{proposition}
\label{prop}
The optimization of $\max_{S(t)} F(S(t))$ subject to the knapsack constraint $\Delta H(S(t)) \leq \Delta D(t)$ can be approximately solved in $O(n\epsilon^{-1}\log\epsilon^{-1})$ time complexity with a $\frac{1}{2}-\epsilon$ approximation ratio guarantee, where $n$ represents the scale of the data.
\end{proposition}

\vspace{-0.1in}

After obtaining the scheduled input step $c_i(t)$ by solving Eq.(\ref{max}), the rationale distillation loss at stage $t$ can be formulated as:
\begin{equation}
\label{lrt}
\setlength{\abovedisplayskip}{0.05in}
\setlength{\belowdisplayskip}{0.05in}
    \mathcal{L}_r (t) = - \sum_i \sum_{j=p_{c_i(t)+1}}^{q_{n_i}} \log P (r_j^{(i)} | r_{<j}^{(i)}, x^{(i)};\theta^s),
\end{equation}
where $p_{c_i(t)+1}$ is the start position of the $(c_i(t)+1)$-th step in the rationale  $r^{(i)}$, and $q_{n_i}$ is the end position of the last step in the rationale $r^{(i)}$.
For each rationale, $n_i$ is fixed and $c_i(t)$ is gradually decreased to $0$. In this way, the student model could learn the rationale of each question in an easy-to-hard manner.


\subsection{Training Procedure}
To train our whole framework, we first optimize the rationale token weighting module by Eq.(\ref{weighting}) to determine the token significance. 
Then, we assess the step difficulty and derive the progressive distillation strategy by solving Eq.(\ref{max}). 
Finally, by integrating these two modules, the overall loss for distilling the rationale at stage $t$ can be written as:
\begin{equation}
\setlength{\abovedisplayskip}{0.05in}
\setlength{\belowdisplayskip}{0.05in}
\label{lot}
    \mathcal{L}_o (t) = - \sum_i \sum_{j=p_{c_i(t)+1}}^{q_{n_i}} w_j^{(i)} \cdot \log P (r_j^{(i)} | r_{<j}^{(i)}, x^{(i)};\theta^s).
\end{equation}
By optimizing $\mathcal{L}_o (t)$, the student model is encouraged to mimic the keypoint tokens precisely, as well as acquiring reasoning capabilities in an easy-to-hard manner. 
Note that we have omitted the inclusion of the prediction loss term for $y^{(i)}$  (referring to the second term in Eq. (\ref{gen})), for the sake of clarity, as it remains constant.
The pseudo-code of our training procedure is listed in Appendix \ref{code}. 

\section{Experiments}

\begin{table*}[]
\centering
\vspace{-0.2in}
\caption{Performance comparison of our method and baselines.}
 \label{overall}
  \setlength{\tabcolsep}{2mm}
 {
\begin{tabular}{@{}ccccccc@{}}
\toprule
\multirow{2}{*}{Models}        & \multirow{2}{*}{\#   Params.} & \multirow{2}{*}{Distillation Methods} & \multicolumn{4}{c}{Datasets}           \\ \cmidrule(l){4-7} 
                              &                               &                                        & GSM8K & ASDiv & SVAMP & CommonsenseQA \\ \midrule
GPT-3.5-Turbo                 & unknown                             & -                                      & 73.98 & 79.64 & 75.14 & 74.35         \\ \midrule
\multirow{6}{*}{LLaMA-7B}     & \multirow{6}{*}{7B}           & -                                      & 11.00      & 40.20      & 32.80      &  33.90             \\
                              &                               & SCoTD                                  & 38.54 & 63.38 & 62.67 & 71.33         \\
                              &                               & Specialized   KD                       & 39.15 & 64.01 & 63.33 & 72.32         \\
                              &                               & SCOTT                                  & 40.97 & 62.74 & 61.33 & 74.45         \\
                              &                               & MCC-KD                                 & 41.58 & 65.76 & 64.67 & 76.41         \\
                              &                               & KPOD   (ours)                          & \textbf{46.74} & \textbf{71.02} & \textbf{68.67} & \textbf{77.89}         \\ \midrule
\multirow{6}{*}{FlanT5-XL}    & \multirow{6}{*}{3B}           & -                                      & 13.50      & 20.70      &  17.70     &  72.70             \\
                              &                               & SCoTD                                  & 21.85 & 25.16 & 26.67 & 79.61         \\
                              &                               & Specialized   KD                       & 23.22 & 28.03 & 25.33 & 81.16         \\
                              &                               & SCOTT                                  & 21.09 & 25.48 & 24.67 & 83.62         \\
                              &                               & MCC-KD                                 & 24.28 & 31.35 & 30.00 & 82.88         \\
                              &                               & KPOD   (ours)                          & \textbf{25.19} & \textbf{33.76} & \textbf{34.67} & \textbf{88.04}         \\ \midrule
\multirow{6}{*}{FlanT5-Large} & \multirow{6}{*}{760M}         & -                                      &  6.90     & 10.10      & 6.80      &  67.60             \\
                              &                               & SCoTD                                  & 19.42 & 20.06 & 19.33 & 76.58              \\
                              &                               & Specialized   KD                       & 20.03 & 23.25 & 20.67 & 77.23              \\
                              &                               & SCOTT                                  & 18.21 & 21.66 & 18.67 & 77.48              \\
                              &                               & MCC-KD                                 & 18.36 & 23.89 & 21.33 & 78.13              \\
                              &                               & KPOD   (ours)                          & \textbf{22.46} & \textbf{27.39} & \textbf{25.33} & \textbf{81.41}              \\ \bottomrule
\end{tabular}
}
\vspace{-0.1in}
\end{table*}

\subsection{Experiment Setup}
In this section, we introduce our experiment settings. The implementation details can be found in Appendix \ref{impl}. 

\textbf{Datasets.} We evaluate our method on both mathematical reasoning tasks and commonsense reasoning tasks, following \cite{hsieh2023distilling,fu2023specializing}. For mathematical reasoning, we adopt three benchmark datasets for evaluation: GSM8K \cite{cobbe2021training}, ASDiv \cite{patel2021nlp} and SVAMP \cite{miao2021diverse}. For commonsense reasoning,  CommonsenseQA benchmark \cite{talmor2019commonsenseqa} is employed to evaluate our method. Additionally, we conduct out-of-distribution (OOD) evaluation via training our method on GSM8K while testing it on ASDiv and SVAMP, following \cite{fu2023specializing}. The dataset splits can be found in Appendix \ref{impl}. 

\textbf{Models and Baselines.} We adopt GPT-3.5-Turbo \cite{ye2023comprehensive} as the teacher model to generate the rationale for each question in the dataset via zero-shot  CoT prompting \cite{kojima2022large}, following \cite{chenetal2023mcc}. This is accessed via the OpenAI's public API for ChatGPT. As for the student model, we adopt three widely-used pretrained language models of different architectures: LLaMA-7B \cite{touvron2023llama}, FlanT5-XL \cite{chung2022scaling} and FlanT5-Large \cite{chung2022scaling}, similar to \cite{fu2023specializing, chenetal2023mcc}. The parameter counts of LLaMA-7B, FlanT5-XL, FlanT5-Large are 7B, 3B, 760M respectively. As for baselines, we employ four state-of-the-art CoT distillation methods for comparison: Specialized KD \cite{fu2023specializing}, SCOTT \cite{wangetal2023scott}, SCoTD \cite{lietal2023symbolic}, MCC-KD \cite{chenetal2023mcc}.
Following previous works \cite{fu2023specializing}, we use the accuracy (\%) metric for evaluating the performance of our method and baselines.

\subsection{Overall Performance}
In this section, we evaluate the overall performance of our method. We compare our method with four recent state-of-the-art CoT distillation methods as mentioned before. The GPT-3.5-Turbo serves as the teacher model. 
Table \ref{overall} illustrates the results.
The symbol \enquote{-}  denotes the model without using CoT distillation methods.
First, we can observe that CoT distillation methods consistently boost the performance of smaller student models on reasoning tasks, underscoring the effectiveness of distilling rationales. 
In addition, it's evident that our proposed KPOD outperforms previous methods by a large margin. For example, compared to MCC-KD, achieving the second best results when using LLaMA-7B as the student model, our approach achieves $5.16\%$, $5.26\%$,  $4.00\%$, $1.48\%$ performance gains on the GSM8K, ASDiv, SVAMP, CommonsenseQA datasets, respectively. This highlights  the effectiveness of promoting precise mimicry of keypoint tokens and implementing a learning schedule that progresses from easy to challenging tasks. Such an approach facilitates the acquisition of reasoning capabilities by the student model.

\begin{table}[]
\centering
\vspace{-0.2in}
\caption{Ablation study of our method.}
 \label{ablation}
  \setlength{\tabcolsep}{0.5mm}
 {
\begin{tabular}{@{}cccc@{}}
\toprule
\multirow{2}{*}{Models}    & \multirow{2}{*}{Settings} & \multicolumn{2}{c}{Datasets} \\ \cmidrule(l){3-4} 
                           &                           & GSM8K     & CommonsenseQA    \\ \midrule
\multirow{8}{*}{LLaMA-7B}  & KPOD-w.o.-sig             & 42.64     & 75.18            \\
                           & KPOD-w.o.-sig-dif         & 44.01     & 76.49            \\
                           & KPOD-w.o.-prog            & 43.25     & 74.61            \\
                           & KPOD-w.o.-div             & 44.16     & 75.76            \\
                           & KPOD-ACL          & 43.55   & 75.51           \\
                           & KPOD-SPL             & 42.94   &  75.84           \\
                           & KPOD-ICL             & 43.85  &  75.35           \\
                           & KPOD                      & 46.74     & 77.89            \\ \midrule
\multirow{8}{*}{FlanT5-XL} & KPOD-w.o.-sig             & 22.46     & 85.26            \\
                           & KPOD-w.o.-sig-dif         & 23.82     & 86.08            \\
                           & KPOD-w.o.-prog            & 23.22     & 84.28            \\
                           & KPOD-w.o.-div             & 23.98     & 86.73            \\
                           & KPOD-ACL             & 23.52   &   86.40         \\
                           & KPOD-SPL             & 22.76   &   85.59         \\
                           & KPOD-ICL             & 22.91   &  85.83           \\
                           & KPOD                      & 25.19     & 88.04            \\ \bottomrule
\end{tabular}
}
\vspace{-0.3in}
\end{table}

\subsection{Ablation Study}

We conduct ablation study to verify the effectiveness of the components in our proposed method. Specifically, we design several variants of our proposed KPOD: KPOD-w.o.-sig denotes our method wherein each token is treated equally, without incorporating the token significance weight for distillation. KPOD-w.o.-sig-dif represents our method without using the  token significance weight for calculating the step difficulty. KPOD-w.o.-prog means our method  without using the proposed progressive distillation strategy. 
KPOD-w.o.-div denotes our method without using the diversity term in the value function to select the question set.

Besides, we compare our method with three representative curriculum learning methods: Adaptive CL \cite{kong2021adaptive}, SPL \cite{wanetal2020self} and ICL \cite{jiaetal2023sample}. We design three variants of our method: KPOD-ACL, KPOD-SPL, KPOD-ICL respectively denote replacing our in-rationale progressive distillation strategy by Adaptive CL, SPL and ICL. The results are listed in Table \ref{ablation}.

As shown in Table \ref{ablation}, KPOD-w.o.-sig obtains inferior performance than KPOD, illustrating the effectiveness of emphasizing the precise mimicry of keypoint tokens in our method. Besides, KPOD outperforms KPOD-w.o.-sig-dif. This shows that it's essential to utilizing the token significance weight for the step difficulty calculation. The performance of KPOD-w.o.-prog is worse than KPOD, illustrating the effectiveness of scheduling an easy-to-hard learning order for CoT distillation. Moreover, KPOD obtains better performance than KPOD-w.o.-div. This demonstrates that ensuring a diverse question set to increase difficulty is effective. Finally, we can find that KPOD surpasses  KPOD-ACL, KPOD-SPL and KPOD-ICL, showing the superiority of our in-rationale progressive distillation strategy compared to previous curriculum learning methods.

\subsection{OOD Performance}
Following \cite{fu2023specializing}, we examine the out-of-distribution (OOD) generalization ability of the student model trained by our method and baselines. We use the in-distribution mathematical dataset GSM8K for training and adopt OOD mathematical datasets ASDiv, SVAMP for testing, similar to \cite{fu2023specializing, chenetal2023mcc}. 
As shown in Table \ref{ood}, our proposed KPOD consistently obtains superior performance compared to the baselines, indicating that the student model trained by our method has stronger  OOD generalization capabilities.

\begin{table}[]
\centering
\vspace{-0.2in}
\caption{OOD performance of our method and baselines.}
 \label{ood}
  \setlength{\tabcolsep}{0.3mm}
 {
\begin{tabular}{@{}ccc|cc@{}}
\toprule
\multirow{2}{*}{Model}     & \multirow{2}{*}{Methods} & In-distribution & \multicolumn{2}{c}{OOD} \\ \cmidrule(l){3-5} 
                           &                                       & GSM8K           & ASDiv              & SVAMP              \\ \midrule
\multirow{5}{*}{LLaMA-7B}  & SCoTD                                 & 38.54           & 55.09                   & 45.33                  \\
                           & Specialized   KD                      & 39.15           & 53.82                   & 38.67                   \\
                           & SCOTT                                 & 40.97           & 53.50                   & 42.00                   \\
                           & MCC-KD                                & 41.58           & 57.64              & 41.00              \\
                           & KPOD   (ours)                         & \textbf{46.74}           & \textbf{57.96}              & \textbf{47.33}              \\ \midrule
\multirow{5}{*}{FlanT5-XL} & SCoTD                                 & 21.85           & 25.48                   & 22.67              \\
                           & Specialized   KD                      & 23.22           & 26.11                   & 24.67                   \\
                           & SCOTT                                 & 21.09           & 25.20                  & 25.33              \\
                           & MCC-KD                                & 24.28           & 28.98              & 26.67              \\
                           & KPOD   (ours)                         & \textbf{25.19}           & \textbf{32.48}              & \textbf{29.33}              \\ \bottomrule
\end{tabular}
}
\vspace{-0.3in}
\end{table}

\subsection{Visualizations}

\begin{figure}
  \centering
  \includegraphics[width=0.95\linewidth]{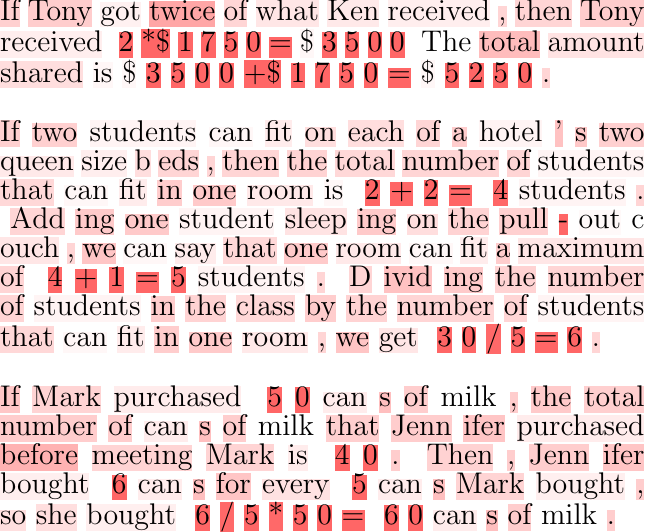}
    \vspace{-0.1in}
  \caption{
Visualizations of token significance weights produced by the weight generator. The intensity of red corresponds to the significance weight assigned to each token, with a deeper red indicating higher weight.
  }
  \label{vis}
  \vspace{-0.1in}
\end{figure}

In this section, we visualize the token significance weight $w_j^{(i)}$ generated by the weight generator, to intuitively show the effectiveness of the rationale token weighting module. 
Figure \ref{vis} illustrates the visualization results on the GSM8K dataset.
First, we can find that the digit tokens and operation tokens obtain the highest weights. 
This is because these tokens are usually of vital importance in the reasoning process, where even a slight deviation could cause errors. 
Additionally, several tokens that contribute significantly to the reasoning also exhibit relatively high weights. Tokens such as \enquote{twice}, \enquote{total}, \enquote{adding}, and \enquote{dividing} provide instructional cues for the reasoning steps. Besides, meaningful subjects like \enquote{Mark} and \enquote{Jennifer} can play a crucial role in reasoning, as their relationships should be considered during the reasoning process. Furthermore, it could be observed that some tokens of less importance for the reasoning are given low weights, such as \enquote{can}, \enquote{say}, \enquote{fit}, \enquote{received}, \enquote{got}, etc. These visualizations demonstrate our rationale token weighting module can effectively determine the significance of rationale tokens, thereby facilitating the student to accurately mimic crucial keypoint tokens. 

\begin{figure}
\centering
\vspace{-0.1in}
\subfigure[performance with varying $\alpha$]{
\begin{minipage}[t]{0.5\linewidth}
\centering
\includegraphics[width=1.5in]{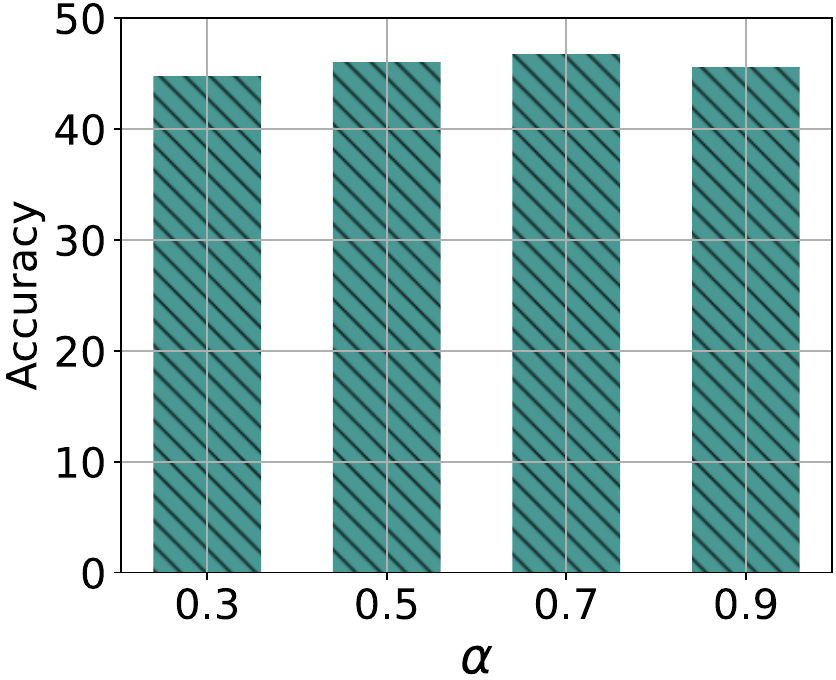}
\end{minipage}%
}%
\subfigure[performance with varying $\beta$]{
\begin{minipage}[t]{0.5\linewidth}
\centering
\includegraphics[width=1.5in]{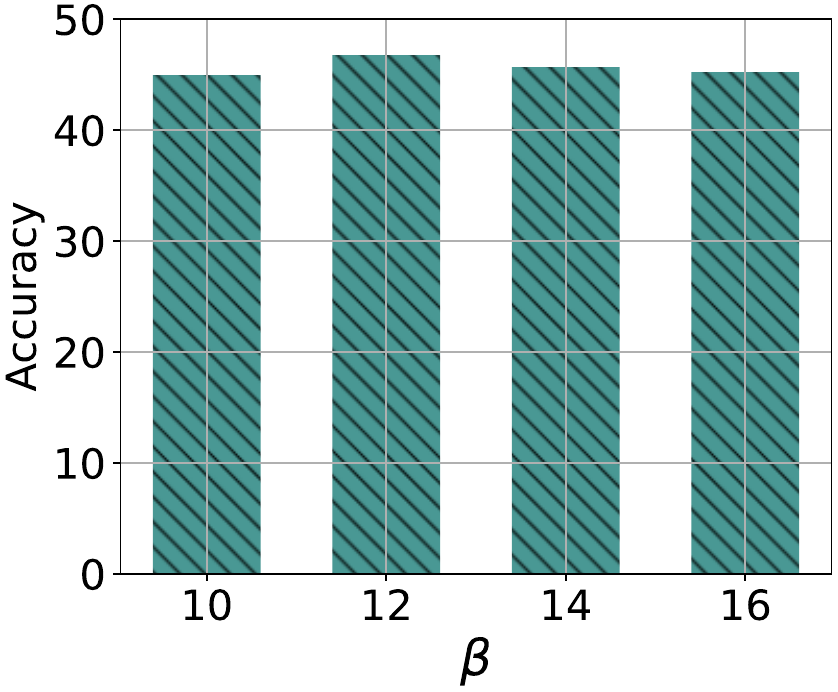}
\end{minipage}%
}%
\centering
\vspace{-0.25in}
\caption{Parameter sensitivity study of $\alpha$ and $\beta$.}
\vspace{-0.3in}
\label{sens1}
\end{figure}

\subsection{Parameter Sensitivity Analysis}
We perform experiments to analyze the effect of two important hyper-parameters $\alpha$ and $\beta$ in our method on GSM8K with LLaMA-7B as the student model. Figure \ref{sens1} shows the results. First, we analyze the effect of hyper-parameter $\alpha$ in the mask ratio  loss. 
 We can observe that the performance of our method is not sensitive to $\alpha$ in a relatively large range. Second, we study the influence of hyper-parameter $\beta$ in the diversity term for question set selection. Similarly,  our method is not sensitive to $\beta$ in a relatively large range. Thus it's easy to set them in practice. We analyze the sensitivity of other hyper-parameters in Appendix \ref{exp}.

\section{Conclusion}
In this paper, we proposed a keypoint-based progressive chain-of-thought distillation framework for LLMs. Specifically, we devised a rationale token weighting module to encourage the student model to accurately mimic keypoint tokens during the distillation process. 
Besides, we proposed an in-rationale progressive distillation strategy to enable the student model to acquire reasoning capabilities from the teacher LLMs in an easy-to-hard manner. Extensive experiments validated the effectiveness of our proposed method.

\section*{Acknowledgment}
This work was supported by the NSFC under Grants 62122013, U2001211. This work was also supported by the Innovative Development Joint Fund Key Projects of Shandong NSF under Grants ZR2022LZH007.

\section*{Impact Statement}
This paper presents work whose goal is to advance the field of machine learning. There are many potential societal consequences of our work, none which we feel must be specifically highlighted here.


\bibliography{example_paper}
\bibliographystyle{icml2024}

\newpage
\appendix
\onecolumn
\section{Implementation Details}
\label{impl}
We perform our experiments using GeForce RTX 3090 GPUs. In order to accelerate training, we employ LoRA \cite{hu2021lora} to train the student model. Following previous CoT distillation works \cite{chenetal2023mcc}, the rank of LoRA is set to 64 for LLaMA-7B and 128 for FlanT5-XL. We use Adam optimizer for optimization with a learning rate of $1\times10^{-5}$ for LLaMA-7B and $5\times10^{-5}$ for FlanT5 models. The batch size is set to 4. 
In terms of LLaMA-7B, the epoch number for training the student model is set to 20 for the GSM8K, CommonsenseQA datasets, and 40 for the ASDiv, SVAMP datasets. 
As for FlanT5 models, the epoch number is set to 100 because they require more optimization steps for convergence.
The input embedding layer in this module aligns with the pretrained student model's input embedding layer for the consistency of tokenizer.
The hyper-parameters $\alpha$ that balances the answer prediction loss and mask ratio loss is set to $0.5$. As for the progressive distillation strategy, we simply treat each epoch as a training state in this paper. The stage $T$ that achieves the maximum difficulty is set as half of the epoch number. 
The initial overall learning difficulty $C_0$ is set to $30\%$ of the maximum difficulty $B$. We set exponential $p=0.5$ in $D(t)$ that controls the growth rate of the learning difficulty. The hyper-parameter $\beta$ of the diversity term in the question set selection is set to 12. The number of clusters is set as $5$ for clustering the question. 

We follow previous CoT distillation works to split the datasets \cite{chenetal2023mcc,fu2023specializing}, the datasets statistics are summarized in Table \ref{dataset}.

\begin{table}[H]
\vspace{-0.2in}
\centering
\caption{Dataset statistics.}
 \label{dataset}
  \setlength{\tabcolsep}{1.5mm}
 {
\begin{tabular}{@{}cccc@{}}
\toprule
Datasets      & Train   Size & Validation   Size & Test   Size \\ \midrule
GSM8K         & 7473         & 660               & 659         \\
ASDiv         & 1462         & 313               & 314         \\
SVAMP         & 700          & 150               & 150         \\
CommonsenseQA & 8520         & 1221              & 1221        \\ \bottomrule
\end{tabular}
}
\end{table}

\section{Training Pseudo-code}
\label{code}
Algorithm 1 outlines the training procedure of our KPOD. Initially, we employ the CoT prompt \cite{kojima2022large} to instruct the teacher LLM to generate step-by-step rationales for each question in the dataset. Subsequently, the rationale token weighting module receives these rationales as input and is trained to determine the significance weights for each token. Following this, we compute the difficulty of each step in the rationale based on these weights. We then utilize the FTGP algorithm \cite{li2022submodular} to maximize Eq.(\ref{max}) to schedule the question set for increasing difficulty at each stage. 
Once scheduled, we train the student model using Eq.(\ref{lot}) based on the established learning order and token significance weights. Before epoch $T$, we progressively escalate the learning difficulty to the maximum difficulty. Post-epoch $T$, the student is trained to generate the complete rationale for each question. This approach allows the student model to precisely mimic the keypoint tokens while progressively acquiring reasoning capabilities in an easy-to-hard fashion.

\begin{algorithm*}
  \setcounter{algorithm}{0}
  \caption{The training procedure of KPOD}  
  \begin{algorithmic}[0]  
    \Require 
    a teacher LLM, dataset $\mathcal{D}=\{(x^{(i)},y^{(i)})\}$, epoch number $N_e$ for training student, epoch number $T$ for achieving the maximum difficulty, hyper-parameter settings;
    \Ensure
    a trained smaller student model $\theta^s$;
    \State prompt the teacher LLM to generate rationale for each question $x^{(i)}$ in $\mathcal{D}$;
    \State optimize the  rationale token weighting module by Eq.(\ref{weighting}) to obtain the token significance weight $w_j^{(i)}$;
    \State calculate the step difficulty based by Eq.(\ref{diff}) based on $w_j^{(i)}$;
    \State run FTGP algorithm \cite{li2022submodular} to solve Eq.(\ref{max}) to derive $S(t)$ for each stage.
    \For{each epoch $e$ from $1$ to $N_e$}
        \State Let $c_i(t)=0$ for every sample;
        \If{$e \leq T$}
            \State obtain $c_i(t)$ by Eq.(\ref{ci}) based on $S(t)$;
        \EndIf
        \State train the student model $\theta^s$ to generate the rationale by Eq.(\ref{lot}) based on $w_j^{(i)}$ and $c_i(t)$;
    \EndFor

  \end{algorithmic}  
  \label{training}
\end{algorithm*}

\section{Additional Experiments}
\label{exp}

\begin{figure*}
\centering
\subfigure[performance with varying $p$]{
\begin{minipage}[t]{0.3\linewidth}
\centering
\includegraphics[width=1.8in]{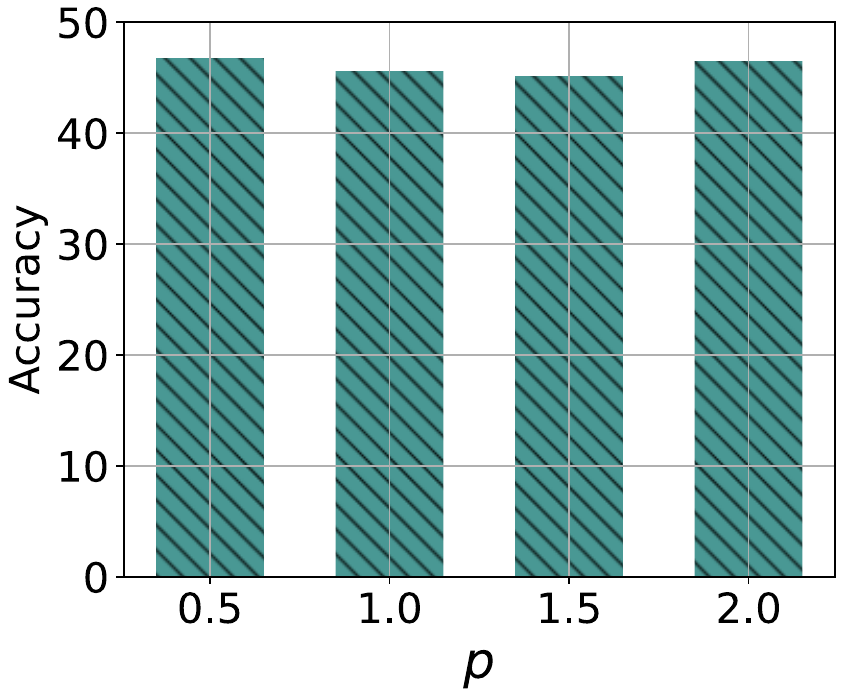}
\end{minipage}%
}%
\subfigure[performance with varying $K$]{
\begin{minipage}[t]{0.3\linewidth}
\centering
\includegraphics[width=1.8in]{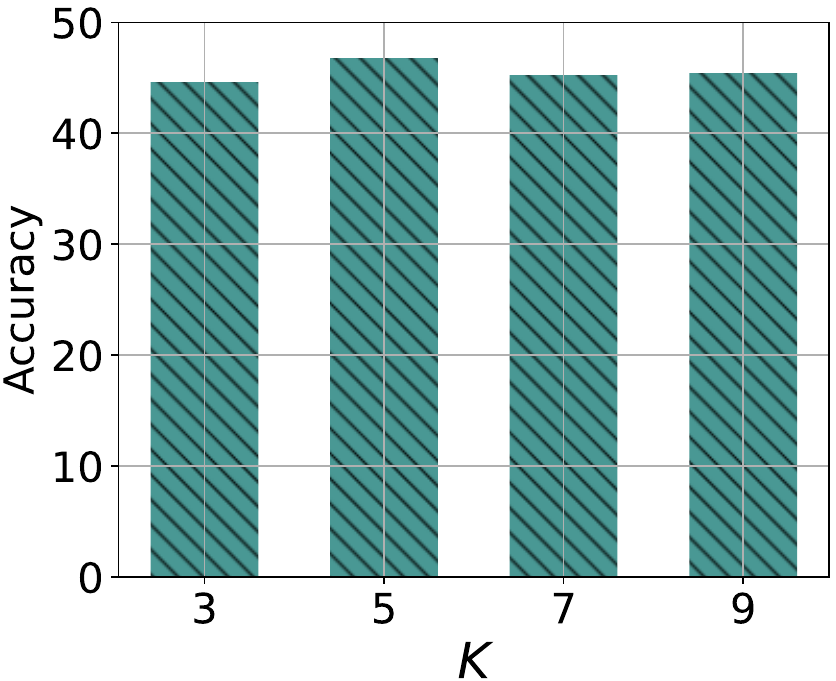}
\end{minipage}%
}%
\centering
\subfigure[performance with varying $C_0$]{
\begin{minipage}[t]{0.3\linewidth}
\centering
\includegraphics[width=1.8in]{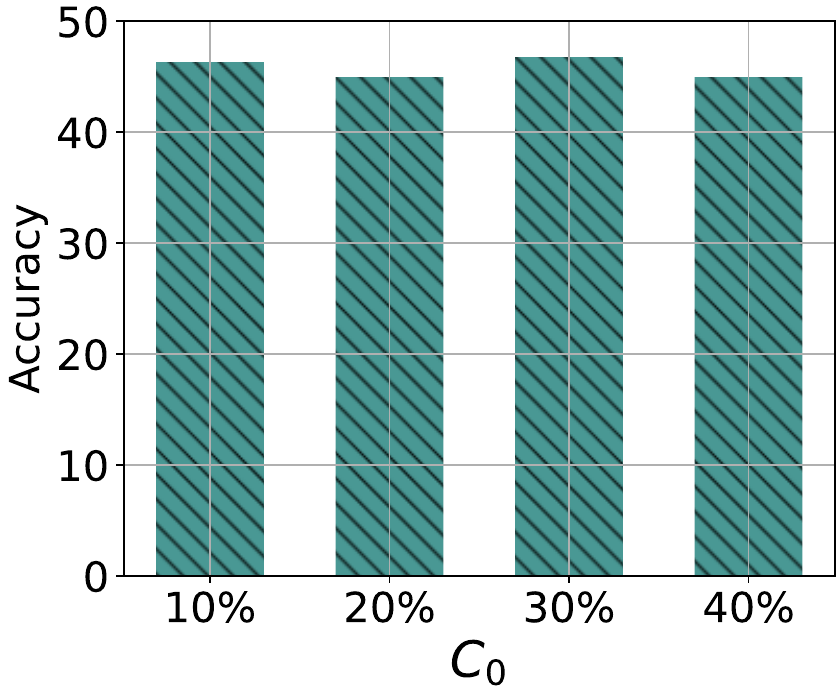}
\end{minipage}%
}%
\centering
\caption{Parameter sensitivity study of $p$, $K$ and $C_0$ on GSM8K.}
\vspace{-0.1in}
\label{sens2}
\end{figure*}

We additionally analyze the sensitivity of three hyper-parameters of $p$, $K$ and $C_0$. Figure 4 (a)(b)(c) show the performance of LLaMA-7B on GSM8K with varying  $p$, $K$, $C_0$ respectively. First, we analyze the effect of $p$ that controls the growth rate the learning difficulty. We can find that the performance of our method is relatively stable to this hyper-parameter. Besides, we study the influence of the number of clusters $K$ for clustering the questions. It can be observed that our method is not sensitive to $K$ in a relatively large range. In addition, we investigate the sensitivity of $C_0$ which is the initial learning difficulty. In Figure 4(c), $r\%$ denotes setting $C_0$ to $r$ percentage of the maximum difficulty $B$. Our method is still not sensitive to this hyper-parameter.

\section{Proof}
\label{proof}
In this section, we prove the Proposition \ref{prop}. First, we introduce Theorem \ref{ftgp} proposed in FTGP algorithm \cite{li2022submodular}.

\begin{theorem}
\label{ftgp}
If a function $f:2^N\rightarrow\mathbb{R}$ is monotone and submodular. 
Then, the optimization of $\max_{S} F(S)$ that subjects to a knapsack constraint can be approximately solved in $O(n\epsilon^{-1}\log\epsilon^{-1})$ time complexity by FTGP  algorithm \cite{li2022submodular} with an approximation ratio guarantee, where $n$ represents the scale of the data and $\epsilon$ is a hyper-parameter.
If ${S^{opt}}$ is the optimal solution and $\hat{S}$ is the approximate solution of FTGP, then $F(\hat{S})\geq(\frac{1}{2}-\epsilon)F(S^{opt})$ holds.
\end{theorem}

According to Theorem \ref{ftgp}, if we could prove that our value function $F$ is monotone and submodular, then Proposition \ref{prop} is proved. In the next, we will prove that our value function $F$ satisfies these two conditions.

\textbf{Definition 1.} (Monotonicity) A function $f:2^N\rightarrow\mathbb{R}$ is monotone if for $\forall A\subseteq B\subseteq N$ where $N$ is the universal set of all elements, it holds that $F(A)\leq F(B) $.

\textbf{Lemma 1.} Our value function $F$ in Eq.(\ref{max}) is monotone.
\begin{proof}
  We define two question sets $A(t),B(t)$ for increasing difficulty at stage $t$ that satisfy $A(t)\subseteq B(t)\subseteq N$. 
  Let $\Delta=F(B(t))-F(A(t))$. We have:
\begin{align}\nonumber
   \Delta &=  -(D(t) - D(t)) + \Delta H(B(t))- \Delta H(A(t)) + \beta \sum_{k=1}^K \sqrt{|C_k \cap B(t)|} - \beta \sum_{k=1}^K \sqrt{|C_k \cap A(t)|} \\ \nonumber 
   & \geq \beta \sum_{k=1}^K \sqrt{|C_k \cap B(t)|} - \beta \sum_{k=1}^K \sqrt{|C_k \cap A(t)|} \\ \nonumber 
   & = \beta \sum_{k=1}^K (\sqrt{|C_k \cap B(t)|} - \sqrt{|C_k \cap A(t)|}) \\ \nonumber 
   & \geq 0
\end{align}

Thus, we have:
\begin{align}
    \Delta&=F(B)-F(A) \geq 0.\\
    &\Rightarrow F(A) \leq F(B).
\end{align}
\end{proof}
\textbf{Definition 2.} (Submodularity) A function $f:2^N\rightarrow\mathbb{R}$ is submodular if for $\forall A\subseteq B\subseteq N$ and $\forall x\in N\backslash B$, it holds that $ F(A\cup\{x\})-F(A)\geq F(B\cup\{x\})-F(B)$.

\textbf{Lemma 2.} Our value function $F$  in Eq.(\ref{max}) is submodular.
\begin{proof}
We define two triad sets $A,B$ that satisfy $A\subseteq B\subseteq N$. Let $T=B\backslash A$. Define $\Delta=(F(A\cup\{x\})-F(A))-(F(B\cup\{x\})-F(B))$. Then we have:
\begin{align}\nonumber
   \Delta &= (\Delta H(A(t)\cup\{x\})- \Delta H(A(t))) - ( \Delta H(B(t)\cup\{x\})- \Delta H(B(t))) \\ \nonumber
   & \ \ \ \ + (\beta \sum_{k=1}^K \sqrt{|C_k \cap (A(t)\cup\{x\})|} - \beta \sum_{k=1}^K \sqrt{|C_k \cap A(t)|}) - (\beta \sum_{k=1}^K \sqrt{|C_k \cap (B(t)\cup\{x\})|} - \beta \sum_{k=1}^K \sqrt{|C_k \cap B(t)|}) \\ \nonumber
   & = \Delta H(\{x\}) - \Delta H(\{x\})  \\ \nonumber
   &\ \ \ \ + (\beta \sum_{k=1}^K \sqrt{|C_k \cap (A(t)\cup\{x\})|} - \beta \sum_{k=1}^K \sqrt{|C_k \cap A(t)|}) - (\beta \sum_{k=1}^K \sqrt{|C_k \cap (B(t)\cup\{x\})|} - \beta \sum_{k=1}^K \sqrt{|C_k \cap B(t)|}) \\ 
   & =  (\beta \sum_{k=1}^K \sqrt{|C_k \cap (A(t)\cup\{x\})|} - \beta \sum_{k=1}^K \sqrt{|C_k \cap A(t)|}) - (\beta \sum_{k=1}^K \sqrt{|C_k \cap (B(t)\cup\{x\})|} - \beta \sum_{k=1}^K \sqrt{|C_k \cap B(t)|}).
\end{align}
Given that $x\in N\backslash B$ and $A \subseteq B $,  it follows that $x \notin A$ and  $x \notin B$. Then, we have:
\begin{align} \nonumber
    \Delta &=  (\beta \sum_{k=1}^K \sqrt{|C_k \cap A(t)|+|C_k\cap\{x\})|} - \beta \sum_{k=1}^K \sqrt{|C_k \cap A(t)|})  \\ 
     &\ \ \ \  - (\beta \sum_{k=1}^K \sqrt{|C_k \cap B(t)| + |C_k\cap\{x\})|} - \beta \sum_{k=1}^K \sqrt{|C_k \cap B(t)|}).
\end{align}

For convenience, we denote $x_k=|C_k \cap A(t)|$, $y_k=|C_k \cap B(t)|$, $z_k=|C_k\cap\{x\})|$. Then, we have:
\begin{align} \nonumber
    \Delta & = \beta \sum_{k=1}^K ((\sqrt{x_k+z_k}-\sqrt{x_k})-(\sqrt{y_k+z_k}-\sqrt{y_k})) \\ \nonumber
    & = \beta \sum_{k=1}^K (\frac{(\sqrt{x_k+z_k}-\sqrt{x_k})-(\sqrt{y_k+z_k}-\sqrt{y_k})(\sqrt{x_k+z_k}+\sqrt{x_k})+(\sqrt{y_k+z_k}+\sqrt{y_k})}{(\sqrt{x_k+z_k}+\sqrt{x_k})+(\sqrt{y_k+z_k}+\sqrt{y_k})}) \\ \nonumber
    & = \beta \sum_{k=1}^K (\frac{x_k+z_k-x_k-(y_k+z_k)+y_k+2\sqrt{x_k+z_k}\sqrt{y_k}-2\sqrt{x_k}\sqrt{y_k+z_k}}{(\sqrt{x_k+z_k}+\sqrt{x_k})+(\sqrt{y_k+z_k}+\sqrt{y_k})}) \\ \nonumber
    & = \beta \sum_{k=1}^K (\frac{2\sqrt{x_k+z_k}\sqrt{y_k}-2\sqrt{x_k}\sqrt{y_k+z_k}}{(\sqrt{x_k+z_k}+\sqrt{x_k})+(\sqrt{y_k+z_k}+\sqrt{y_k})}) \\ \nonumber
    & = \beta \sum_{k=1}^K (\frac{2\sqrt{x_ky_k+y_kz_k}-2\sqrt{x_ky_k+x_kz_k}}{(\sqrt{x_k+z_k}+\sqrt{x_k})+(\sqrt{y_k+z_k}+\sqrt{y_k})}) \\ \nonumber
\end{align}
Since $A \subseteq B$, it's evident that $y_k = |C_k \cap B(t)| \geq |C_k \cap A(t)| = x_k$. Therefore, we conclude: 
\begin{align} \nonumber
    \Delta 
    & = \beta \sum_{k=1}^K (\frac{2\sqrt{x_ky_k+y_kz_k}-2\sqrt{x_ky_k+x_kz_k}}{(\sqrt{x_k+z_k}+\sqrt{x_k})+(\sqrt{y_k+z_k}+\sqrt{y_k})}) \\ \nonumber
    & \geq \beta \sum_{k=1}^K (\frac{2\sqrt{x_ky_k+x_kz_k}-2\sqrt{x_ky_k+x_kz_k}}{(\sqrt{x_k+z_k}+\sqrt{x_k})+(\sqrt{y_k+z_k}+\sqrt{y_k})}) = 0
\end{align}

Then, we can derive:
\begin{align}
    \Delta&=(F(A\cup\{x\})-F(A))-(F(B\cup\{x\})-F(B)) \geq 0. \\ 
    &\Rightarrow F(A\cup\{x\})-F(A) \geq F(B\cup\{x\})-F(B).
\end{align}
\end{proof}

\end{document}